\documentclass{article}

\usepackage{microtype}
\usepackage{graphicx}
\usepackage{subfigure}
\usepackage{booktabs} 

\usepackage{hyperref}
\usepackage{url}

\usepackage[accepted]{icml2019}

\usepackage{amsmath,amssymb,amsthm}
\usepackage{mathtools}
\usepackage{enumitem}

\usepackage{algorithm}
\usepackage{algorithmic}
\usepackage{multirow}

\newcommand{\eps}{\epsilon}

\newcommand{\R}{\mathbb{R}}

\DeclarePairedDelimiter{\norm}{\lVert}{\rVert}

\newcommand{\ourAE}{$\ell_1$-AE }

\newcommand{\relu}{\textnormal{ReLU}}
\newcommand{\supp}{\textnormal{supp}}
\newcommand{\sign}{\textnormal{sign}}
\newcommand{\nullsp}{\textnormal{null}}

\newtheorem{lemma}{Lemma}

\newtheorem{definition}{Definition}

\newtheorem*{app-lemma}{Lemma}
\newtheorem*{app-theorem}{Theorem}
\newtheorem*{app-definition}{Definition}

\begin{document}

\twocolumn[
\icmltitle{Learning a Compressed Sensing Measurement Matrix via Gradient Unrolling}

\begin{icmlauthorlist}
\icmlauthor{Shanshan Wu}{ut}
\icmlauthor{Alexandros G.~Dimakis}{ut}
\icmlauthor{Sujay Sanghavi}{ut}
\icmlauthor{Felix X. Yu}{goo}
\icmlauthor{Daniel Holtmann-Rice}{goo}
\icmlauthor{Dmitry Storcheus}{goo}
\icmlauthor{Afshin Rostamizadeh}{goo}
\icmlauthor{Sanjiv Kumar}{goo}
\end{icmlauthorlist}

\icmlaffiliation{ut}{Department of Electrical and Computer Engineering, University of Texas at Austin, USA}
\icmlaffiliation{goo}{Google Research, New York, USA}

\icmlcorrespondingauthor{Shanshan Wu}{shanshan@utexas.edu}

\icmlkeywords{Compressed sensing, autoencoder, unroll}

\vskip 0.3in
]

\printAffiliationsAndNotice{}  

\begin{abstract}
Linear encoding of sparse vectors is widely popular, but is commonly data-independent -- missing any possible extra (but a priori unknown) structure beyond sparsity. In this paper we present a new method to learn linear encoders that adapt to data, while still performing well with the widely used $\ell_1$ decoder. The convex $\ell_1$ decoder prevents gradient propagation as needed in standard gradient-based training. Our method is based on the insight that unrolling the convex decoder into $T$ projected subgradient steps can address this issue. Our method can be seen as a data-driven way to learn a compressed sensing measurement matrix. We compare the empirical performance of 10 algorithms over 6 sparse datasets (3 synthetic and 3 real). Our experiments show that there is indeed additional structure beyond sparsity in the real datasets; our method is able to discover it and exploit it to create excellent reconstructions with fewer measurements (by a factor of 1.1-3x) compared to the previous state-of-the-art methods. We illustrate an application of our method in learning label embeddings for extreme multi-label classification, and empirically show that our method is able to match or outperform the precision scores of SLEEC, which is one of the state-of-the-art embedding-based approaches.
\end{abstract}

\section{Introduction}

Assume we have some unknown data vector $x \in \R^d$.
We can observe only a few ($m<d$) linear equations of its entries
and we would like to design these projections by 
creating a measurement matrix $A\in\R^{m\times d}$
such that the projections $y= Ax$ 
allow exact (or near-exact) recovery of the original vector $x\in\R^d$.

If $d>m$, this is an ill-posed problem in general: we are observing $m$ linear equations with $d$ unknowns, so any vector $x'$ on the subspace $A x'=y$ satisfies our observed measurements. 
In this high-dimensional regime, the only hope is to make a structural assumption on $x$, so that unique reconstruction is possible. A natural approach is to assume that the data vector is sparse. The problem of designing measurement matrices and reconstruction algorithms that recover sparse vectors from linear observations is called Compressed Sensing (CS), Sparse Approximation or Sparse Recovery Theory~\cite{Don06, CRT06}.

A natural way to recover is to search for the sparsest solution 
that satisfies the linear measurements: 
\begin{equation}
\vspace{-0.1cm}
\arg \min_{x'\in\R^d}\norm{x'}_0\quad \textnormal{s.t. } Ax'=y. \label{intro-l0min}
\vspace{-0.1cm}
\end{equation}
Unfortunately this problem is NP-hard and for that reason the 
$\ell_0$ norm is relaxed into an $\ell_1$-norm minimization\footnote{Other examples are greedy~\cite{TG07}, and iterative algorithms, e.g., CoSaMP~\cite{NT09}, IHT~\cite{BD09}, and AMP~\cite{DMM09}.}
\begin{equation}
\vspace{-0.1cm}
D(A, y) := \arg \min_{x'\in\R^d}\norm{x'}_1\quad \textnormal{s.t. } Ax'=y. \label{intro-l1min}
\vspace{-0.1cm}
\end{equation}
Remarkably, if the measurement matrix $A$ satisfies some properties, e.g. Restricted Isometry Property (RIP)~\cite{Can08} or the nullspace condition (NSP)~\cite{Rau10}, it is possible to prove 
that the $\ell_1$ minimization in (\ref{intro-l1min}) produces the same output as the intractable $\ell_0$ minimization in (\ref{intro-l0min}). However, it is NP-hard to test whether a matrix satisfies RIP~\cite{BDMS13}.

In this paper we are interested in vectors that are not only sparse but have {\em additional structure} in their support. 
This extra structure may not be known or a-priori specified. We propose a data-driven algorithm that {\em learns a good linear measurement matrix $A$ from data samples}.
Our linear measurements are subsequently decoded with 
the $\ell_1$-minimization in (\ref{intro-l1min}) to estimate the unknown vector $x$.

Many real-world sparse datasets have additional structures beyond simple sparsity. For example, in a demographic dataset with (one-hot encoded) categorical features, a person's income level may be related to his/her education. Similarly, in a text dataset with bag-of-words representation, it is much more likely for two related words (e.g., football and game) to appear in the same document than two unrelated words (e.g., football and biology). A third example is that in a genome dataset, certain groups of gene features may be correlated. In this paper, the goal is to {\em learn} a measurement matrix $A$ to leverage such additional structure. 

Our method is an autoencoder for sparse data, with a linear encoder (the measurement matrix) and a complex non-linear decoder that approximately solves an optimization problem. The latent code is the measurement $y \in \R^m$ which forms the bottleneck of the autoencoder that makes the representation interesting. 
A popular data-driven dimensionality reduction method is Principal Components Analysis (PCA) (see  e.g., \citealt{Hot33,BGKL15,WBSD16,LWZ17}).
PCA is also an autoencoder where both the encoder and decoder are linear and learned from samples. 
Given a data matrix $X\in\R^{n\times d}$ (each row is a sample),
PCA projects each data point $x\in\R^d$ onto the subspace spanned by the top right-singular vectors of $X$. As is well-known, PCA provides the lowest mean-squared error when used with a linear decoder. However, when the data is sparse, non-linear recovery algorithms like (\ref{intro-l1min}) can yield significantly better recovery performance.

In this paper, we focus on learning a linear encoder for sparse data. Compared to non-linear embedding methods such as kernel PCA~\cite{MSSMSR99}, a linear method enjoys two broad advantages: 1) it is easy to compute, as it only needs a matrix-vector multiplication; 2) it is easy to interpret, as every column of the encoding matrix can be viewed as a feature embedding. Interestingly, \citet{AKSV18} recently observed that the pre-trained word embeddings such as GloVe and word2vec~\cite{MSCCD13,PSM14} formed a good measurement matrix for text data. Those measurement matrices, when used with $\ell_1$-minimization, need fewer measurements than the random matrices to achieve near-perfect recovery.

Given a sparse dataset that has additional (but unknown) structure, our goal is to learn a good measurement matrix $A$, when the recovery algorithm is the $\ell_1$-minimization in (\ref{intro-l1min}).
More formally, given $n$ sparse samples $x_1, x_2, \cdots, x_n \in\R^d$, our problem of finding the best $A$ can be formulated as 
\vspace{-0.2cm}
\begin{equation}
\min_{A\in\R^{m\times d} } f(A),\text{ where } f(A) :=\sum_{i=1}^n  \norm{x_i - D(A, Ax_i)}_2^2. \nonumber 
\vspace{-0.2cm}
\end{equation}
Here $D(\cdot, \cdot)$ is the $\ell_1$ decoder defined in (\ref{intro-l1min}). Unfortunately, there is no easy way to compute the gradient of $f(A)$ with respect to $A$, due to the optimization problem defining $D(\cdot, \cdot)$. Our main insight, which will be elaborated on in Section~\ref{sec-intuition}, is that {\bf replacing the $\ell_1$-minimization with a $T$-step projected subgradient update of it}, results in gradients being (approximately) computable. In particular, consider the approximate function $\tilde{f}(A): \R^{m\times d} \mapsto \R$ defined as 
\vspace{-0.2cm}
\begin{equation}
\begin{aligned}
&&\tilde{f}(A) := &\quad \sum_{i=1}^n \norm{x_i - \hat{x}_i}_2^2, \quad \text{ where}\\
&&\hat{x}_i = &\quad \text{$T$-step projected subgradient of }\\
&& &\quad D(A, Ax_i), \text{ for }i=1,\cdots,n. 
\end{aligned}\label{intro-approx-goal}
\end{equation}
As we will show, now it is {\em possible} to compute the gradients of $\tilde{f}(A)$ with respect to $A$. This idea is sometimes called \textit{unrolling} and has appeared in various other applications as we discuss in the related work section. To the best of our knowledge, we are the first to use unrolling to learn a measurement matrix for compressed sensing. 

Our contributions can be summarized as follows:
\vspace{-0.4cm}
\begin{itemize}[leftmargin=*]
\item We design a novel autoencoder, called \ourAE, to learn an efficient and compressed representation for structured sparse vectors. Our autoencoder is easy to train. It has only two tuning hyper-parameters in the network architecture: the encoding dimension $m$ and the network depth $T$. The architecture of \ourAE is inspired by the projected subgradient method of solving the $\ell_1$ decoder in (\ref{intro-l1min}). While the exact projected subgradient method requires computing the pseudoinverse, we circumvent this by observing that it is possible to replace the expensive pseudoinverse operation by a simple transpose (Lemma~\ref{lemma-equal}).  
\item The most surprising result in this paper is that we can learn a linear encoder using an unfolded $T$-step projected subgradient decoder and the learned measurement matrix {\em still} performs well for the original $\ell_1$-minimization decoder. We empirically compare 10 algorithms over 6 sparse datasets (3 synthetic and 3 real). As shown in Figure~\ref{main_plots}, using the measurement matrix learned from our autoencoder, we can compress the sparse vectors (in the test set) to a lower dimension (by a factor of 1.1-3x) than random Gaussian matrices while still being able to {\em perfectly} recover the original sparse vectors (see also Table~\ref{powerful_l1}). This demonstrates the superior ability of our autoencoder in learning and adapting to the additional structures in the given data.
\item Although our autoencoder is specifically designed for $\ell_1$-minimization decoder, the learned measurement matrix also performs well (and can perform even better) with the model-based decoder~\cite{BCDH10} (Figure~\ref{ae_cosamp}). This further demonstrates the benefit of learning a measurement matrix from data. As a baseline algorithm, we slightly modify the original model-based CoSaMP algorithm by adding a positivity constraint without changing the theoretical guarantee (Appendix~\ref{app-cosamp-pos}), which could be of independent interest.
\item Besides the application in compressed sensing, one interesting direction for future research is to use the proposed \ourAE in other supervised tasks. We illustrate a potential application of \ourAE in extreme multi-label learning. We show that \ourAE can be used to learn label embeddings for multi-label datasets. We compare the resulted method with one of the state-of-the-art embedding-based methods SLEEC~\cite{BJKVJ15} over two benchmark datasets. Our method is able to achieve better or comparable precision scores than SLEEC (see Table~\ref{xml}).
\end{itemize}


\section{Related Work}
We briefly review the relevant work, and highlight the differences compared to our paper.

{\bf Model-based compressed sensing (CS).} Model-based CS~\cite{BCDH10, HIS14} extends the conventional compressed sensing theory by considering more realistic structured models than simple sparsity. It requires to know the sparsity structure a priori, which is not always possible in practice. Our approach, by contrast, does not require a priori knowledge about the sparsity structure. 

{\bf Learning-based measurement design.} Most theoretical results in CS are based on random measurement matrices. There are a few approaches proposed to learn a measurement matrix from training data. One approach is to learn a near-isometric embedding that preserves pairwise distance~\cite{HSYB15, BSC13}. This approach usually requires computing the pairwise difference vectors, and hence is computationally expensive if both the number of training samples and the dimension are large (which is the setting that we are interested in). Another approach restricts the form of the measurement matrix, e.g., matrices formed by a subset of rows of a given basis matrix. The learning problem then becomes selecting the best subset of rows~\cite{BLSGBC16, LC16, GMLICSC18}. In Figure~\ref{main_plots}, we compare our method with the learning-based subsampling method proposed in~\cite{BLSGBC16}, and show that our method needs fewer measurements to recover the original sparse vector.

{\bf Adaptive CS.} In adaptive CS~\cite{SN11, ACD13, MN14}, the new measurement is designed based on the previous measurements in order to maximize the gain of new information. This is in contrast to our setting, where we are given a set of training samples. Our goal is to learn a good measurement matrix to leverage additional structure in the given samples.

{\bf Dictionary learning.} Dictionary learning~\cite{AEB06, MBPS09} is the problem of learning an overcomplete set of basis vectors from data so that every datapoint (presumably dense) can be represented as a sparse linear combination of the basis vectors. By contrast, this paper focuses on learning a good measurement matrix for data that are already sparse in the canonical basis.

{\bf Sparse coding.} The goal of sparse coding~\cite{OF96, DE03} is to find the sparsest representation of a dense vector, given a fixed family of basis vectors (aka a dictionary). Training a deep neural network to compute the sparse codes becomes popular recently~\cite{GL10, SBS15, WLH16}. Several recent papers~\cite{KLTKA16, XWGWW16, SJZZ17, JMFU17, MMPVDP17, MPB15, MDB17, MDB19, MB17, HXIW17, ZG17, LKKTA18} proposes new convolutional architectures for image reconstruction from low-dimensional measurements. Some of the networks also have an image sensing component. 
Sparse coding is different from this paper, where our focus is on learning a good measurement/encoding matrix rather than learning a good recovery/decoding algorithm. 

{\bf Autoencoders.} An autoencoder is a popular neural network architecture for unsupervised learning. It has applications in dimensionality reduction~\cite{HS06}, pre-training~\cite{EBCMVB10}, image compression and recovery~\cite{LKKTA18, MPB15, MDB17, MDB19}, denoising~\cite{VLLBM10}, and generative modeling~\cite{KW14}. In this paper we design a novel autoencoder \ourAE to learn a compressed sensing measurement matrix for the $\ell_1$-minimization decoder. 

{\bf Unrolling.} The idea of unfolding an iterative algorithm into a neural network is a natural way to make the algorithm trainable~\cite{GL10, HRW14, SBS15, XWGWW16, WLH16, SJZZ17, JMFU17, MMPVDP17, HXIW17, ZG17, MSDPMVP18, KBRW19}. The main difference between the previous papers and this paper is that, most previous papers seek a trained neural network as a replacement of the original optimization-based algorithm, while in this paper we design an autoencoder based on the unrolling idea, and after training, we show that the learned measurement matrix still performs well with the \emph{original} $\ell_1$-minimization decoder.

{\bf Extreme multi-label learning (XML).} The goal of XML is to learn a classifier to identify (for each datapoint) a subset of relevant labels from a extreme large label set. Different approaches have been proposed for XML, e.g., embedding-based~\cite{BJKVJ15, YJKD14, MK15, Tag17}, tree-based~\cite{PV14, JPV16, JDBPKH16, PKGDHAV18}, one-vs-all~\cite{PKHAV18, BS17, YHRZD16, YHDRDX17, NA17, HVV12}, and deep learning~\cite{JCS17, LCWY17}. Here we focus on the embedding-based approach. In Section~\ref{sec-xml} we show that the proposed autoencoder can be used to learn label embeddings for multi-label datasets.

\section{Our Algorithm}
Our goal is to learn a measurement matrix $A$ from the given sparse vectors. This is done via training a novel autoencoder, called $\ell_1$-AE. In this section, we will describe the key idea behind the design of $\ell_1$-AE. In this paper we focus on the vectors that are sparse in the standard basis and also non-negative\footnote{Extending our method to general cases is left for future work.}. This is a natural setting for many real-world datasets, e.g., categorical data and bag-of-words data.

\subsection{Intuition}\label{sec-intuition}
Our design is strongly motivated by the projected subgradient method used to solve an $\ell_1$-minimization problem. Consider the following $\ell_1$-minimization problem:
\begin{equation}
\min_{x'\in\R^d}\norm{x'}_1\quad \textnormal{s.t. } Ax'=y, \label{intuition-l1min}
\end{equation}
where $A\in\R^{m\times d}$ and $y\in\R^m$. We assume that $m<d$ and that $A$ has rank $m$. One approach\footnote{Another approach is via linear programming.} to solving (\ref{intuition-l1min}) is the projected subgradient method. The update is given by
\begin{equation}
x^{(t+1)} = \Pi(x^{(t)}-\alpha_t g^{(t)}), \text{ where }g^{(t)} = \sign(x^{(t)})\label{intuition-update}
\end{equation}
where $\Pi$ denotes the (Euclidean) projection onto the convex set $\{x': Ax'=y\}$, $g^{(t)}$
is the sign function, i.e., the subgradient of the objective function $\norm{\cdot}_1$ at $x^{(t)}$, and $\alpha_t$ is the step size at the $t$-th iteration. Since $A$ has full row rank, $\Pi$ has a closed-form solution given by
\begin{align}
\Pi(z) &= \arg\;\min_h \norm{h-z}_2^2\quad \textnormal{s.t. } Ah=y \\
& = z + \arg\;\min_{h'} \norm{h'}_2^2\quad \textnormal{s.t. } Ah'=y-Az\\
& = z + A^{\dagger}(y-Az), \label{intuition-proj}
\end{align}
where $A^{\dagger} = A^T(AA^T)^{-1}$ is the Moore-Penrose inverse of matrix $A$. Substituting (\ref{intuition-proj}) into (\ref{intuition-update}), and using the fact that $Ax^{(t)}=y$, we get the following update equation
\begin{equation}
x^{(t+1)} = x^{(t)} -\alpha_t (I - A^{\dagger} A)\sign(x^{(t)}),\label{intuition-final}
\end{equation}
where $I$ is the identity matrix. We use $x^{(1)} = A^{\dagger}y$ (which satisfies the constraint $Ax'=y$) as the starting point.

As mentioned in the Introduction, our main idea is to replace the solution of an $\ell_1$ decoder given in (\ref{intuition-l1min}) by a $T$-step projected subgradient update given in (\ref{intuition-final}). One technical difficulty\footnote{One approach is to replace $A^{\dagger}$ by a trainable matrix $B\in\R^{d\times m}$. This approach performs worse than ours (see Figure~\ref{amazon_var}).} in simulating (\ref{intuition-final}) using neural networks is backpropagating through the pesudoinverse $A^{\dagger}$. Fortunately, Lemma \ref{lemma-equal} shows that it is possible to replace $A^{\dagger}$ by $A^T$. 
\begin{lemma}
For any vector $x\in\R^d$, and any matrix $A\in\R^{m\times d}$ ($m<d$) with rank $m$, there exists an $\tilde{A}\in \R^{m\times d}$ with all singular values being ones, such that the following two $\ell_1$-minimization problems have the same solution:
\begin{align}
&P_1:\quad \min_{x'\in\R^d}\norm{x'}_1\quad \textnormal{s.t. } Ax'=Ax.\\ &P_2:\quad \min_{x'\in\R^d}\norm{x'}_1\quad \textnormal{s.t. } \tilde{A}x'=\tilde{A}x.
\end{align}
Furthermore, the projected subgradient update of $P_2$ is
\begin{equation}
x^{(t+1)} = x^{(t)} - \alpha_t(I-\tilde{A}^T\tilde{A})\sign(x^{(t)}), \quad x^{(1)} = \tilde{A}^T\tilde{A}x.
\label{lemma2-update}
\end{equation}
A natural choice for $\tilde{A}$ is $U(AA^T)^{-1/2}A$, where $U\in\R^{m\times m}$ can be any unitary matrix.
\label{lemma-equal}
\end{lemma}
Lemma 1 essentially says that: 1) Instead of searching over {\em all} matrices (of size $m$-by-$d$ with rank $m$), it is enough to search over a subset of matrices $\tilde{A}$, whose singular values are all ones. This is because $A$ and $\tilde{A}$ has the same recovery performance for $\ell_1$-minimization (this is true as long as $\tilde{A}$ and $A$ have the same null space). 2) The key benefit of searching over matrices with singular values being all ones is that the corresponding projected subgradient update has a simpler form: the annoying pseudoinverse term $A^{\dagger}$ in (\ref{intuition-final}) is replaced by a simple matrix transpose $A^T$ in (\ref{lemma2-update}). 

As we will show in the next section, our decoder is designed to simulate (\ref{lemma2-update}) instead of (\ref{intuition-final}): the only difference is that the pseudoinverse term $A^{\dagger}$ is replaced by matrix transpose $A^T$. Ideally we should train our \ourAE by enforcing the constraint that the matrices have singular values being ones. In practice, we do not enforce that constraint during training\footnote{Efficient training with this manifold constraint~\cite{MJKKM18} is an interesting direction for future work. We empirically found that adding a regularizer $\norm{I-AA^T}_2$ to the loss function would degrade the performance.}. We empirically observe that the learned measurement matrix $A$ is not far from the constraint set (see Appendix~\ref{app-singular}). 

\subsection{Network Structure of \ourAE}

\begin{figure*}[ht]
\centering
\vspace{-0.2cm}
\includegraphics[width=0.86\textwidth]{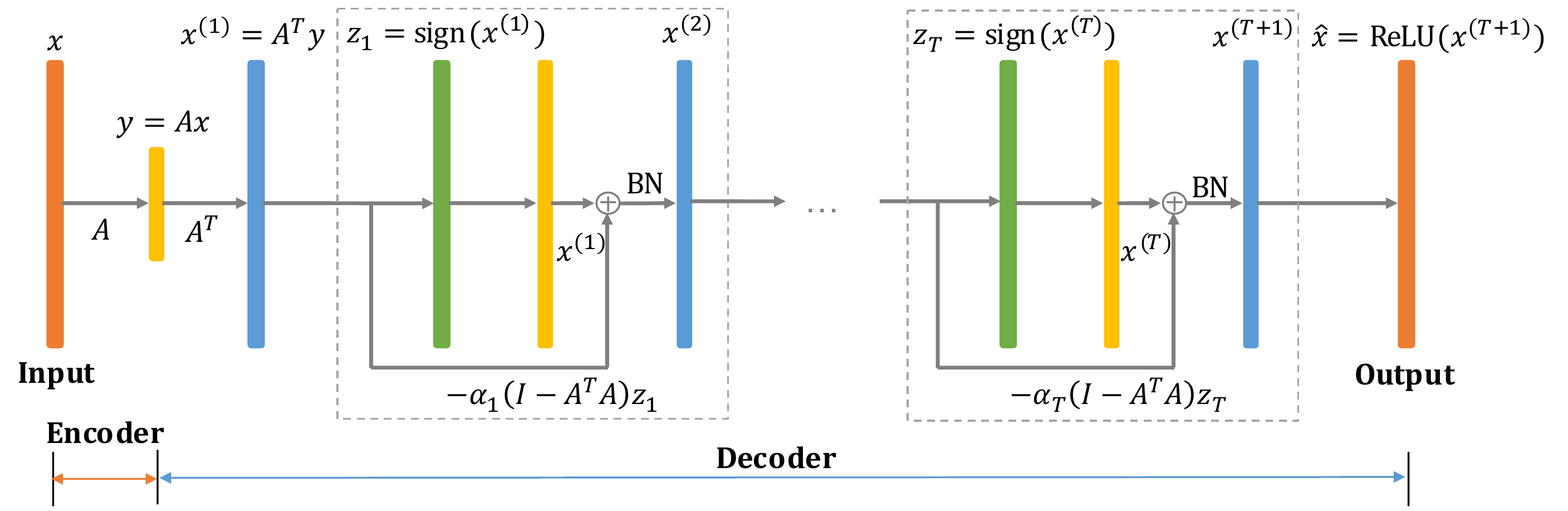}
\vspace{-0.4cm}
\caption{Network structure of the proposed autoencoder \ourAE.}
\label{structure}
\vspace{-0.4cm}
\end{figure*}

As shown in Figure~\ref{structure}, \ourAE has a simple linear encoder and a non-linear decoder. When a data point $x \in\R^d$ comes, it is encoded as $y=Ax$, where $A\in \R^{m\times d}$ is the encoding matrix that will be learned from data. A decoder is then used to recover the original vector $x$ from its embedding $y$. 

The decoder network of \ourAE consists of $T$ blocks connected in a feedforward manner: the output vector of the $t$-th block is the input vector to the $(t+1)$-th block. The network structure inside each block is identical. Let $x^{(1)} = A^Ty$. For $t\in\{1,2,...,T\}$, if $x^{(t)}\in\R^d$ is the input to the $t$-th block, then its output vector $x^{(t+1)}\in\R^d$ is
\vspace{-0.1cm}
\begin{equation}
x^{(t+1)} = x^{(t)} -\alpha_t(I - A^TA)\sign(x^{(t)}),\label{block}
\vspace{-0.1cm}
\end{equation}
where $\alpha_1, \alpha_2, \cdots, \alpha_T\in\R$ are scalar variables to be learned from data. We empirically observe that regularizing $\alpha_t$ to have the following form\footnote{It is called {\em square summable but not summable}~\cite{SM14}.} $\alpha_t = \beta/t$ for $t\in\{1,2,...,T\}$ improves test accuracy. Here, $\beta\in\R$ is the only scalar variable to be learned from data. We also add a standard batch normalization (BN) layer~\cite{IS15} inside each block, because empirically it improves the test accuracy (see Figure~\ref{amazon_var}). After $T$ blocks, we use rectified linear units (ReLU) in the last layer\footnote{This makes sense as we focus on non-negative sparse vectors in this paper. Non-negativity is a natural setting for many real-world sparse datasets, e.g., categorical data and text data.} to obtain the final output $\hat{x}\in\R^d$: $\hat{x} = \relu(x^{(T+1)})$.

It is worth noting that the low-rank structure of the weight matrix $I - A^TA$ in (\ref{block}) is essential for reducing the computational complexity. A fully-connected layer requires a weight matrix of size $d\times d$, which is intractable for large $d$.

Given $n$ unlabeled training examples $\{x_i\}_{i=1}^n$, we will train an \ourAE to minimize the average squared $\ell_2$ distance between $x\in\R^d$ and $\hat{x}\in\R^d$:
\vspace{-0.1cm}
\begin{equation}
\min_{A\in\R^{m\times d},\; \beta\in\R} \quad \frac{1}{n}\sum_{i=1}^n\norm{x_i-\hat{x}_i}_2^2.
\vspace{-0.1cm}
\end{equation}

\section{Experiments}
We implement \ourAE in Tensorflow. Our code is available online: \url{https://github.com/wushanshan/L1AE}. In this section, we demonstrate that \ourAE is able to learn a good measurement matrix $A$ for the structured sparse datasets, when decoding is done by $\ell_1$-minimization\footnote{We use Gurobi (a commercial optimization solver) to solve it.}.

\subsection{Datasets and Training}\label{sec-data}
\begin{table*}[ht]
\small
\centering
\vspace{-0.5cm}
\caption{Summary of the datasets. The validation set is used for parameter tuning and early stopping.}
\begin{tabular}{|c|c|c|c|c|}
\hline
Dataset & Dimension & Avg. no. of nonzeros & Train / Valid / Test Size & Description\\
\hline
Synthetic1 & 1000 & 10 & 6000 / 2000 / 2000 & 1-block sparse with block size 10 \\
Synthetic2 & 1000 & 10 & 6000 / 2000 / 2000 & 2-block sparse with block size 5\\
Synthetic3 & 1000 & 10 & 6000 / 2000 / 2000 & Power-law structured sparsity\\
Amazon & 15626 & 9 & 19661 / 6554 / 6554 & 1-hot encoded categorical data\\
Wiki10-31K & 30398 & 19 & 14146 / 3308 / 3308 & Extreme multi-label data\\
RCV1 & 47236 & 76 & 13889 / 4630 / 4630 & Text data with TF-IDF features\\
\hline
\end{tabular}
\label{datasets}
\vspace{-0.35cm}
\end{table*}
\begin{figure*}[ht]
\centering
\includegraphics[width=0.8\textwidth]{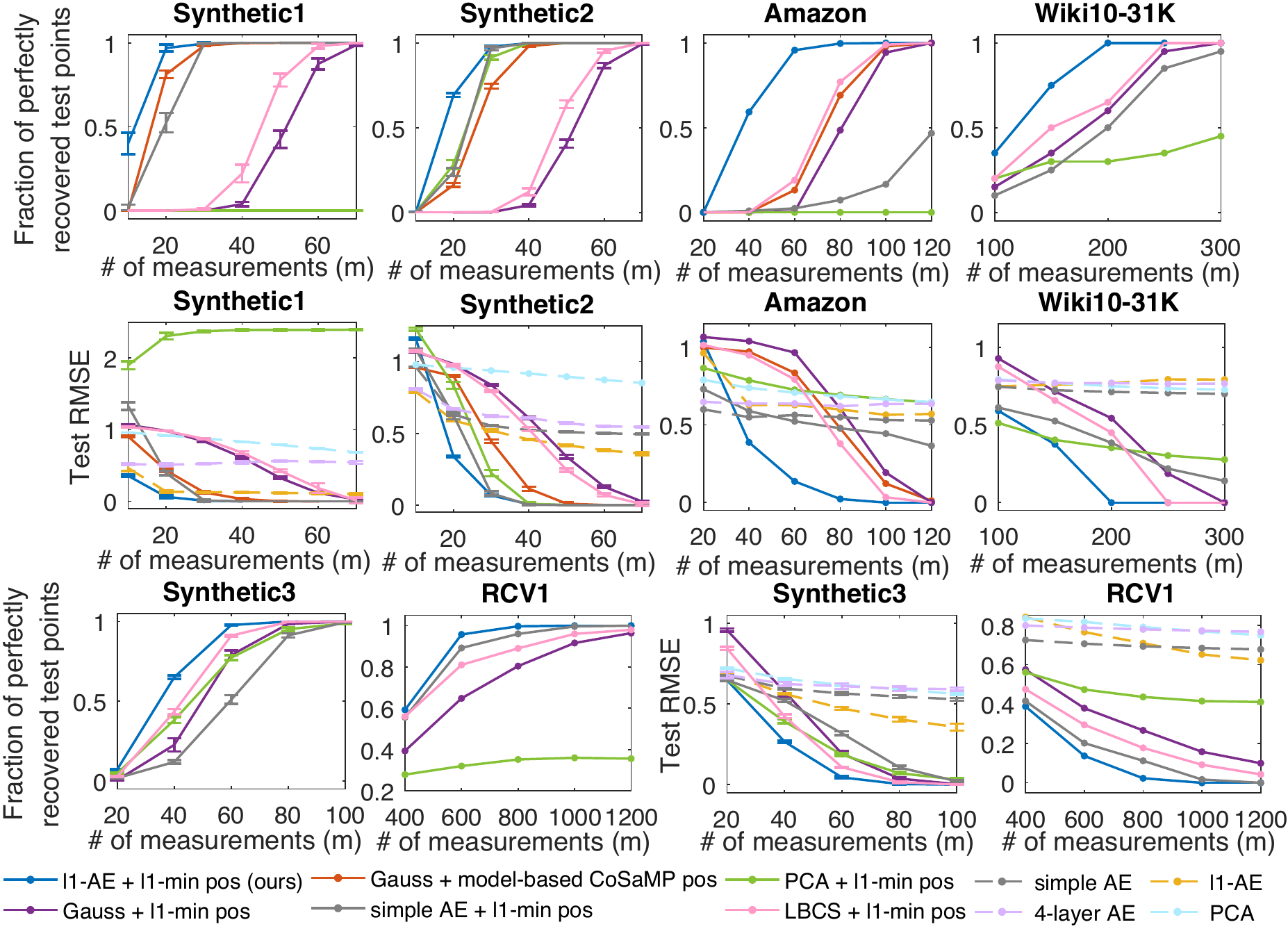}
\vspace{-0.4cm}
\caption{Best viewed in color. Recovery performance over the test set: fraction of exactly recovered data points (the 1st row and the left half of the 3rd row); reconstruction error (the 2nd row and the right half of the 3rd row). For the following four baselines, we do not plot the fraction of recovered points: simple AE, 4-layer AE, $\ell_1$-AE, and PCA. This is because they cannot produce a perfect reconstruction $\hat{x}$ that satisfies $\norm{x-\hat{x}}_2\le 10^{-10}$ (see also Table~\ref{powerful_l1}). For synthetic data, we plot the mean and standard deviation (indicated by the error bars) across 10 randomly generated datasets. Note that model-based CoSaMP decoder is not applicable for the Synthetic3, Wiki10-31K, and the RCV1 datasets. Although the model-based CoSaMP decoder has more information about the given data (such as block sparsity and one-hot sparsity) than the $\ell_1$-minimization decoder, our ``$\ell_1$-AE + $\ell_1$-min pos'' still gives the best recovery performance.}
\label{main_plots}
\vspace{-0.4cm}
\end{figure*}

{\bf Synthetic datasets.} As shown in Table~\ref{datasets}, we generate three synthetic datasets\footnote{We also have experiments on a synthetic dataset with no extra structure (see Appendix~\ref{app-no-extra}).}: two satisfy the block sparsity model\footnote{A signal $x\in\R^d$ is called $K$-block sparse with block size $J$ if it satisfies: 1) $x$ can be reshaped into a matrix $X$ of size $J\times N$, where $JN=d$; 2) every column of $X$ acts as a group, i.e., the entire column is either zero or nonzero; 3) $X$ has $K$ nonzero columns, and hence $x$ has sparsity $KJ$.}~\cite{BCDH10}, and one follows the power-law structured sparsity (The $i$-th coordinate is nonzero with probability proportional to $1/i$).
Each sample is generated as follows: 1) choose a random support set satisfying the sparsity model; 2) set the nonzeros to be uniform in $[0,1]$. 

{\bf Real datasets.} Our first dataset is from Kaggle ``Amazon Employee Access Challenge''\footnote{\url{https://www.kaggle.com/c/amazon-employee-access-challenge}}. Each training example contains 9 categorical features. We use one-hot encoding to transform each example. Our second dataset Wiki10-31K is a multi-label dataset downloaded from this repository~\cite{BDJPV17}. We only use the label vectors to train our autoencoder. Our third dataset is RCV1\cite{LYRL04}, a popular text dataset. We use scikit-learn to fetch the training set and  randomly split it into train/validation/test sets.

{\bf Training.} We use stochastic gradient descent to train the autoencoder. Before training, every sample is normalized to have unit $\ell_2$ norm. The parameters are initialized as follows: $A\in\R^{m\times d}$ is a random Gaussian matrix with standard deviation $1/\sqrt{d}$; $\beta$ is initialized as 1.0. Other hyper-parameters are given in Appendix~\ref{app-train-params}. A single NVIDIA Quadro P5000 GPU is used in the experiments. We set the decoder depth $T=10$ for most datasets\footnote{Although the subgradient method~\cite{SM14} has $1/\eps^2$ convergence, in practice, we found that a small value of $T$ (e.g., $T=10$ in Table~\ref{depth_AE}) was good enough~\cite{GEBS18}.}. Training an \ourAE can be done in several minutes for small-scale datasets and in around an hour for large-scale datasets.

\subsection{Algorithms}\label{sec-algo}

We compare 10 algorithms in terms of their recovery performance. The results are shown in Figure~\ref{main_plots}. All the algorithms follow a two-step ``encoding $+$ decoding'' process. 

{\bf \ourAE + $\ell_1$-min pos (our algorithm)}: After training an \ourAE, we use the encoder matrix $A$ as the measurement matrix. To decode, we use Gurobi (a commercial optimization solver) to solve the following $\ell_1$-minimization problem with positivity constraint (denoted as ``$\ell_1$-min pos''):
\vspace{-0.2cm}
\begin{equation} 
\min_{x'\in\R^d}\norm{x'}_1\quad \textnormal{s.t. } Ax'=y, \;x'\ge 0. \label{l1-pos}
\end{equation}
Since we focus on non-negative sparse vectors in this paper, adding a positivity constraint improves the recovery performance\footnote{The sufficient and necessary condition (Theorem 3.1 of~\citealt{KDXH11}) for exact recovery using (\ref{l1-pos}) is weaker than the nullspace property (NSP)~\cite{Rau10} for (\ref{intuition-l1min}).} (see Appendix~\ref{app-extra-exp-pos-l1}).

{\bf Gauss + $\ell_1$-min pos / model-based CoSaMP pos}: A random Gaussian matrix $G\in\R^{m\times d}$ with i.i.d. $\mathcal{N}(0, 1/m)$ entries is used as the measurement matrix\footnote{Additional experiments with random partial Fourier matrices~\cite{HR15} can be found in Appendix~\ref{app-extra-exp-fourier}.}. We experiment with two decoding methods: 1) Solve the optimization problem given in (\ref{l1-pos}); 2) Use the model-based CoSaMP algorithm\footnote{Model-based method needs the explicit sparsity model, and hence is not applicable for RCV1, Wiki10-31K, and Synthetic3.} (Algorithm 1 in~\citealt{BCDH10}) with an additional positivity constraint (see Appendix~\ref{app-cosamp-pos}). 

{\bf PCA or PCA + $\ell_1$-min pos}: We perform truncated singular value decomposition (SVD) on the training set. Let $A\in\R^{m\times d}$ be the top-$m$ singular vectors. For PCA, every vector $x\in\R^d$ in the test set is estimated as $A^TAx$. We can also use ``$\ell_1$-min pos'' (\ref{l1-pos}) as the decoder.

{\bf Simple AE or Simple AE + $\ell_1$-min pos}: We train a simple autoencoder: for an input vector $x\in\R^d$, the output is $\relu(B^TAx) \in\R^d$, where both $B\in\R^{m\times d}$ and $A\in\R^{m\times d}$ are learned from data. We use the same loss function as our autoencoder. After training, we use the learned $A$ matrix as the measurement matrix. Decoding is performed either by the learned decoder or solving ``$\ell_1$-min pos'' (\ref{l1-pos}).

{\bf LBCS + $\ell_1$-min pos / model-based CoSaMP pos}: We implement the learning-based compressive subsampling (LBCS) method in~\cite{BLSGBC16}. The idea is to select a subset (of size $m$) of coordinates (in the transformed space) that preserves the most energy. We use Gaussian matrix as the basis matrix and ``$\ell_1$-min pos'' as the decoder\footnote{We tried four variations of LBCS: two different basis matrices (random Gaussian matrix and DCT matrix), two different decoders ($\ell_1$-minimization and linear decoder). The combination of Gaussian and $\ell_1$-minimization performs the best (see Appendix~\ref{app-LBCS}).}. Decoding with ``model-based CoSaMP pos'' is in Figure~\ref{ae_cosamp}.

{\bf 4-layer AE}: We train a standard 4-layer autoencoder (we do not count the input layer), whose encoder network (and decoder network) consists of two feedforward layers with ReLU activation. The dimension of the 1st (and 3rd) layer is tuned based on the performance on the validation set. 

\subsection{Results and Analysis}\label{sec-res}
\begin{figure*}[ht]
\centering
\vspace{-0.2cm}
\begin{subfigure}	
    \centering
	\includegraphics[width=0.53\textwidth]{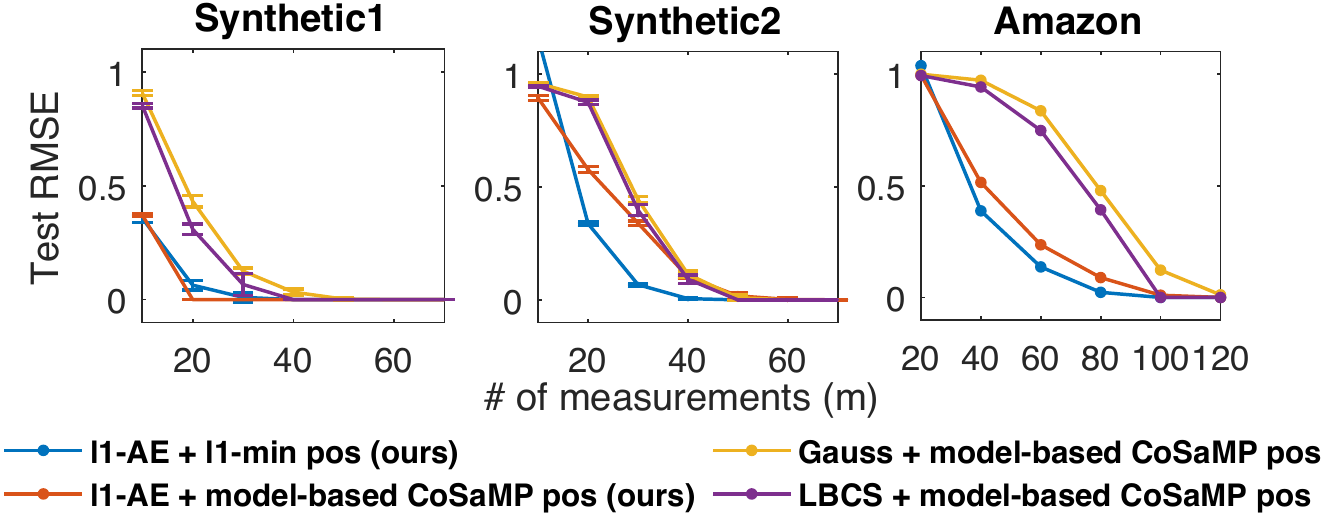}
\end{subfigure}
\begin{subfigure}
    \centering
    \includegraphics[width=0.32\textwidth]{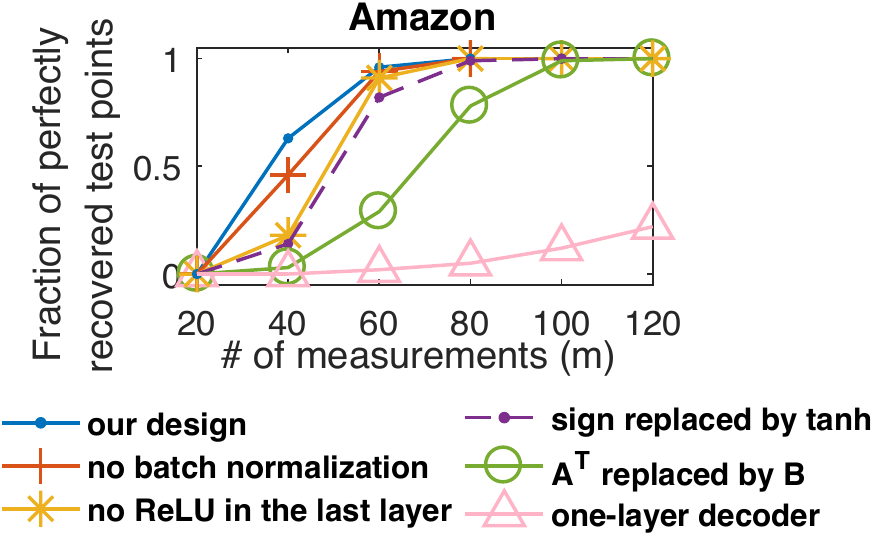}
\end{subfigure}
\vspace{-0.4cm}
\caption{Best viewed in color. {\bf Left three plots}: Although \ourAE is designed for the $\ell_1$-min decoder, the matrix learned from \ourAE also improves the recovery performance when the decoder is model-based CoSaMP. {\bf Right plot}: Recovery performance of ``\ourAE + $\ell_1$-min pos'' on the Amazon test set when we slightly change the decoder structure. Each change is applied in isolation.}
\label{ae_cosamp}
\label{amazon_var}
\vspace{-0.4cm}
\end{figure*}

The experimental results are shown in Figure~\ref{main_plots}. Two performance metrics are compared. The first one is the fraction of exactly recovered test samples. We say that a vector $x$ is exactly recovered by an algorithm if it produces an $\hat{x}$ that satisfies $\norm{x-\hat{x}}_2\le 10^{-10}$. The second metric is the root mean-squared error (RMSE) over the test set\footnote{Note that in Figure~\ref{main_plots}, test RMSE has similar scale across all datasets, because the vectors are normalized to have unit $\ell_2$ norm.}.


As shown in Figure~\ref{main_plots}, our algorithm ``\ourAE + $\ell_1$-min pos'' outperforms the rest baselines across all datasets. By learning a data-dependent linear encoding matrix, our method requires fewer measurements to achieve perfect recovery.

{\bf Learned decoder versus $\ell_1$-min decoder.} We now compare two methods: $\ell_1$-AE and $\ell_1$-AE + $\ell_1$-min pos. They have a common encoder but different decoders. As shown in Figure~\ref{main_plots}, ``$\ell_1$-AE + $\ell_1$-min pos" almost always gives a smaller RMSE. In fact, as shown in Table~\ref{powerful_l1}, ``$\ell_1$-min pos" is able to achieve reconstruction errors in the order of 1e-15, which is impossible for a neural network. The strength of optimization-based decoder over neural network-based decoder has been observed before, e.g., see Figure 1 in \cite{BJPD17}\footnote{As indicated by Figure 1 in \cite{BJPD17}, LASSO gives better reconstruction than GAN-based method when given enough Gaussian measurements.}. Nevertheless, neural network-based decoder usually has an advantage that it can handle nonlinear encoders, for which the corresponding optimization problem may become non-convex and hard to solve exactly.

\begin{table*}[ht]
\small
\centering
\vspace{-0.15cm}
\caption{{\bf Left}: Comparison of test RMSE: $\ell_1$-AE versus $\ell_1$-AE + $\ell_1$-min pos. {\bf Right}: Test RMSE of our method "$\ell_1$-AE + $\ell_1$-min pos" on the synthetic1 dataset: the error decreases as the decoder depth $T$ increases.}
\begin{tabular}{|c|c|c|c|c|c|c|}
\hline
Dataset & \multicolumn{3}{|c|}{Synthetic1}&  \multicolumn{3}{|c|}{Amazon}\\
\hline
\# measurements & 10 & 30  & 50  & 40 & 80 & 120  \\
 \hline
$\ell_1$-AE & 0.465 & 0.129  & 0.118 &  0.638  & 0.599 & 0.565\\
\hline
$\ell_1$-AE + $\ell_1$-min pos (ours) & \textbf{0.357} & \textbf{9.9e-3} & \textbf{1.4e-15} &
\textbf{0.387}  & \textbf{0.023} & \textbf{2.8e-15}\\
\hline
\end{tabular}
\begin{tabular}{|c|c|c|c|}
\hline
\# measurements  & 20 & 30\\
\hline
$T=10$ & 0.097 & 9.9e-3\\
$T=20$ & 0.063 & 1.2e-15 \\
$T=30$ & 9.6e-16 & 1.1e-15 \\
\hline
\end{tabular}
\label{depth_AE}
\label{powerful_l1}
\vspace{-0.5cm}
\end{table*}

{\bf Model-based decoder versus learned encoder.} It is interesting to see that our algorithm even outperforms model-based method~\cite{BCDH10}, even though the model-based decoder has more information about the given data than $\ell_1$-minimization decoder. For the Amazon dataset, compared to ``Gauss + model-based CoSaMP pos'', our method reduces the number of measurements needed for exact recovery by about 2x. This is possible because the model-based decoder only knows that the input vector comes from one-hot encoding, which is a {\em coarse} model for the underlying sparsity model. By contrast, \ourAE learns a measurement matrix directly from the given training data.

A natural question to ask is whether the measurement matrix learned by \ourAE can improve the recovery performance of the model-based decoding algorithm. As shown in Figure~\ref{ae_cosamp}, the recovery performance of ``\ourAE + model-based CoSaMP pos'' is better than ``Gauss + model-based CoSaMP pos''. This further demonstrates the benefit of learning a data-adaptive measurement matrix. 

{\bf Variations of $\ell_1$-AE.} We now examine how slightly varying the decoder structure would affect the performance. We make the following changes to the decoder structure: 1) remove the Batch Normalization layer; 2) remove the ReLU operation in the last layer; 3) change the nonlinearity from $\sign$ to $\tanh$; 4) replace the $A^T$ term in the decoder network by a matrix $B\in\R^{d\times m}$ that is learned from data; 5) use one-layer neural network as the decoder, i.e., set $T=0$ in $\ell_1$-AE. Each change is applied in isolation. As shown in Figure~\ref{amazon_var}, our design gives the best recovery performance among all the variations for the Amazon dataset\footnote{Besides these variations, we can design new autoencoders by unrolling other decoding algorithms such as ISTA~\cite{GL10}. More discussions can be found in Appendix~\ref{app-LISTA-AE}.}.

{\bf Decoder depth of $\ell_1$-AE.} The decoder depth $T$ is a tuning parameter of \ourAE. Empirically, the performance of the learned matrix improves as $T$ increases (see Table~\ref{depth_AE}). On the other hand, the training time increases as $T$ increases. The parameter $T$ is tuned as follows: we start with $T=5, 10, ..., T_{\max}$, and stop if the validation performance improvement is smaller than a threshold or if $T$ equals $T_{\max}$. The hyper-parameters used for training \ourAE are given in Appendix~\ref{app-train-params}. We set $T=5$ for Synthetic2 and Synthetic3, $T=10$ for Synthetic1, RCV1, and Wiki10-31K, and $T=60$ for Amazon dataset. The autoencoder is trained on a single GPU. Training an \ourAE takes a few minutes for small datasets and around an hour for large datasets.

\section{Extreme Multi-label Learning}\label{sec-xml}
We have proposed a novel autoencoder \ourAE to learn a compressed sensing measurement matrix for high-dimensional sparse data. Besides the application in compressed sensing, one interesting direction for future research is to use \ourAE in other machine learning tasks. Here we illustrate a potential application of \ourAE in extreme multi-label learning (XML). For every input feature vector, the goal of XML is to predict a subset of relevant labels from a extremely large label set. As a result, the output label vector is high-dimensional, sparse, and non-negative (with 1's denoting the relevant labels and 0's otherwise).

Many approaches have been proposed for XML~\cite{BDJPV17}. Here we focus on the embedding-based approach, and one of the state-of-the-art embedding-based approaches\footnote{AnnexML~\cite{Tag17} is a graph-embedding approach for XML. Some of its techniques (such as better partition of the input data points) can be potentially used with our method.} is SLEEC~\cite{BJKVJ15}. Given $n$ training samples $(x_i, y_i)$, $i=1,...,n$, where $x_i\in\R^p$, $y_i\in\R^d$, we use $X\in\R^{p\times n}$ and $Y\in\R^{d\times n}$ to denote the stacked feature matrix and label matrix. SLEEC works in two steps. In Step 1, SLEEC reduces the dimension of the labels $Y$ by learning a low-dimensional embedding for the training samples. Let $Z\in\R^{m\times n}$ ($m<d$) be the learned embedding matrix. Note that only the $Y$ matrix is used for learning $Z$ in this step. In Step 2, SLEEC learns a linear mapping $V\in\R^{m\times p}$ such that $Z\approx VX$. To predict the label vector for a new sample $x\in\R^{p}$, SLEEC uses nearest neighbors method: first computes the embedding $Vx$, identifies a few nearest neighbors (from the training set) in the embedding space, and uses the sum of their label vectors as prediction.  

The method that we propose follows SLEEC's two-step procedure. The main difference is that in Step 1, we train an autoencoder \ourAE to learn embeddings for the labels $Y$. Note that in XML, the label vectors $Y$ are high-dimensional, sparse, and non-negative. Let $A\in\R^{m\times d}$ be the learned measurement matrix for $Y$, then the embedding matrix is $Z=AY$. In Step 2, we use the same subroutine as SLEEC to learn a linear mapping from $X$ to $Z$. To predict the label vector for a new sample, we compared three methods in our experiments: 1) use the nearest neighbor method (same as SLEEC); 2) use the decoder of the trained \ourAE (which maps from the embedding space to label space); 3) use an average of the label vectors obtained from 1) and 2). The three methods are denoted as ``\ourAE  1/2/3'' in Table~\ref{xml}.

In Table~\ref{xml}, we compare the precision score P@1 over two benchmark datasets. The 2nd row is the number of models combined in the ensemble. According to~\cite{BDJPV17}, SLEEC achieves a precision score 0.7926 for EURLex-4K and 0.8588 for Wiki10-31K, which are consistent with what we obtained by running their code. The embedding dimensions are $m=100$ for EURLex-4K and $m=75$ for Wiki10-31K. We set $T=10$ for $\ell_1$-AE. As shown in Table~\ref{xml}, our method has higher score than SLEEC for EURLex-4K. For Wiki10-31K, a single model of our method has higher score than SLEEC. When 3 or 5 models are ensembled, our method has comparable precision score to SLEEC. More results can be found in Appendix~\ref{app-xml}.

\begin{table}[ht]
\small
\centering
\vspace{-0.5cm}
\caption{{\bf Upper}: Description of the datasets. {\bf Lower}: Comparisons of P@1 scores. The 2nd row is the no. of models in the ensemble.}
\begin{tabular}{|p{1.62cm}|p{1.1cm}|p{1.1cm}|p{1.3cm}|p{0.95cm}|}
\hline
\multirow{2}{*}{Dataset} & Feature & Label & Train/ & \# labels \\ 
& dimension & dimension &Test & /point\\
\hline
EURLex-4K & 5000 & 3993 & 15539/3809 & 5.31 \\
Wiki10-31K & 101938& 30938& 14146/6616 & 18.64\\
\hline
\end{tabular}
\begin{tabular}{|p{1.2cm}|p{0.68cm}|p{0.68cm}|p{0.68cm}|p{0.68cm}|p{0.68cm}|p{0.68cm}|}
\hline
Dataset & \multicolumn{3}{|c|}{EURLex-4K} & \multicolumn{3}{|c|}{Wiki10-31K} \\
\hline
\# models & 1&3&5&1&3&5\\
\hline
SLEEC & 0.7600 & 0.7900&0.7944 & 0.8356 & 0.8603 & 0.8600 \\
\ourAE 1 & 0.7655&0.7928&0.7931 & 0.8529&0.8564&0.8597\\
\ourAE 2 & 0.7949&0.8033&0.8070 & 0.8560&0.8579&0.8581\\
\ourAE 3 & {\bf 0.8062}& {\bf 0.8151}&{\bf 0.8136}&{\bf 0.8617}&{\bf 0.8640}&{\bf 0.8630}\\
\hline
\end{tabular}
\label{xml}
\vspace{-0.4cm}
\end{table}

\section{Conclusion}

Combining ideas from compressed sensing, convex optimization and deep learning, we proposed a novel unsupervised learning framework for high-dimensional sparse data.
The proposed autoencoder \ourAE is able to learn an efficient measurement matrix by adapting to the sparsity structure of the given data. The learned measurement matrices can be subsequently used in other machine learning tasks such as extreme multi-label learning. 
We expect that the learned \ourAE can lead to useful representations in various supervised learning pipelines, for datasets that are well represented by large sparse vectors.
Investigating the relation between the training data and the learned matrix (see Appendix~\ref{app-toy} for a toy example) is an interesting direction for future research. 

\section{Acknowledgements}
This research has been supported by NSF Grants 1302435, 1564000, and 1618689, DMS 1723052, CCF 1763702, ARO YIP W911NF-14-1-0258 and research gifts by Google, Western Digital and NVIDIA.

\bibliography{ref}
\bibliographystyle{icml2019}

\clearpage
\newpage
\appendix
\section{Proof of Lemma~\ref{lemma-equal}}
For convenience, we re-state Lemma~\ref{lemma-equal} here and then give the proof.
\vspace{1em}
\begin{app-lemma}
For any vector $x\in\R^d$, and any matrix $A\in\R^{m\times d}$ ($m<d$) with rank $m$, there exists an $\tilde{A}\in \R^{m\times d}$ with all singular values being ones, such that the following two $\ell_1$-norm minimization problems have the same solution:
\begin{align}
&P_1:\quad \min_{x'\in\R^d}\norm{x'}_1\quad \textnormal{s.t. } Ax'=Ax.\\ &P_2:\quad \min_{x'\in\R^d}\norm{x'}_1\quad \textnormal{s.t. } \tilde{A}x'=\tilde{A}x.
\end{align}
Furthermore, the projected subgradient update of $P_2$ is given as
\begin{equation}
x^{(t+1)} = x^{(t)} - \alpha_t(I-\tilde{A}^T\tilde{A})\sign(x^{(t)}), \quad x^{(1)} = \tilde{A}^T\tilde{A}x. \nonumber
\end{equation}
A natural choice for $\tilde{A}$ is $U(AA^T)^{-1/2}A$, where $U\in\R^{m\times m}$ can be any unitary matrix.
\end{app-lemma}

\begin{proof}
To prove that $P_1$ and $P_2$ give the same solution, it suffices to show that their constraint sets are equal, i.e.,
\begin{equation}
\{x: Ax=Az\} = \{x: \tilde{A}x = \tilde{A}z\}. 
\end{equation}
Since $\{x: Ax=Az\} = \{z+v: v\in\nullsp(A)\}$ and $\{x: \tilde{A}x = \tilde{A}z\} = \{z+v: v\in\nullsp(\tilde{A})\}$, it then suffices to show that $A$ and $\tilde{A}$ have the same nullspace:
\begin{equation}
\nullsp(A) = \nullsp(\tilde{A}). \label{app-null}
\end{equation}
If $v$ satisfies $Av=0$, then $U(AA^T)^{-1/2}Av=0$, which implies $\tilde{A}v=0$. Conversely, we suppose that $\tilde{A}v=0$. Since $U$ is unitary, $AA^T\in\R^{m\times m}$ is full-rank, $(AA^T)^{(1/2)}U^T\tilde{A}v=0$, which implies that $Av=0$. Therefore, (\ref{app-null}) holds.

The projected subgradient of $P_2$ has the following update
\begin{align}
x^{(t+1)} &= x^{(t)} - \alpha_t(I-\tilde{A}^T(\tilde{A}\tilde{A}^T)^{-1}\tilde{A})\sign(x^{(t)}), \\
x^{(1)} &= \tilde{A}^T(\tilde{A}\tilde{A}^T)^{-1}\tilde{A}z\label{tildeA-update}
\end{align}
Since $\tilde{A} = U(AA^T)^{-1/2}A$, we have 
\begin{align}
\tilde{A}\tilde{A}^T &=  U(AA^T)^{-1/2}AA^T (AA^T)^{-1/2} U^T \nonumber\\
&=  U(AA^T)^{-1/2}(AA^T)^{1/2}(AA^T)^{1/2}(AA^T)^{-1/2}U^T \nonumber\\
&= I. \label{tildeA-identity}
\end{align}
Substituting (\ref{tildeA-identity}) into (\ref{tildeA-update}) gives the desired recursion:
\begin{equation}
x^{(t+1)} = x^{(t)} - \alpha_t(I-\tilde{A}^T\tilde{A})\sign(x^{(t)}),\quad x^{(1)} = \tilde{A}^T\tilde{A}z.\nonumber
\end{equation}
\end{proof}

\section{Training parameters}\label{app-train-params}

\begin{table*}[ht]
\centering
\begin{tabular}{|c|c|c|c|c|c|c|}
\hline
Dataset & Depth & Batch size & Learning rate & $N_{\max}$ & $N_{\text{validation}}$ & $N_{\text{no improve}}$\\
\hline
Toy &  10 & 128 & 0.01 & 2e4 & 10 & 5\\
Synthetic1 &  10 & 128 & 0.01 & 2e4 & 10 & 5\\
Synthetic2 & 5 & 128 & 0.01 & 2e4 & 10 & 1\\
Synthetic3 & 5 & 128 & 0.01 & 2e4 & 10 & 1\\
Amazon &  60 & 256 & 0.01 & 2e4 & 1 & 1\\
Wiki10-31K & 10 & 256 & 0.001 & 5e3 & 10 & 1\\
RCV1 & 10 & 256 & 0.001 & 1e3 & 1 & 50\\
\hline
\end{tabular}
\caption{Training parameters.}
\label{train_params}
\end{table*}

Table~\ref{train_params} lists the parameters used to train \ourAE in our experiments. We explain the parameters as follows.
\begin{itemize}
\item Depth: The number of blocks in the decoder, indicated by $T$ in Figure~\ref{structure}.
\item Batch size: The number of training samples in a batch.
\item Learning rate: The learning rate for SGD.
\item $N_{\max}$: Maximum number of training epochs.
\item $N_{\text{validation}}$: Validation error is computed every $N_{\text{validation}}$ epochs. This is used for early-stopping.
\item $N_{\text{no improve}}$: Training is stopped if the validation error does not improve for $N_{\text{no improve}}*N_{\text{validation}}$ epochs.
\end{itemize}

\section{Model-based CoSaMP with additional positivity constraint} \label{app-cosamp-pos}
The CoSaMP algorithm~\cite{NT09} is a simple iterative and greedy algorithm used to recover a $K$-sparse vector from the linear measurements. The model-based CoSaMP algorithm (Algorithm 1 of~\cite{BCDH10}) is a modification of the CoSaMP algorithm. It uses the prior knowledge about the support of the $K$-sparse vector, which is assumed to follow a predefined {\em structured sparsity model}. In this section we slightly modify the model-based CoSaMP algorithm to ensure that the output vector follows the given sparsity model and is also {\em nonnegative}. 

To present the pseudocode, we need a few definitions. We begin with a formal definition for the structured sparsity model $\mathcal{M}_K$ and the sparse approximation algorithm $\mathbb{M}$. For a vector $x\in\R^d$, let $x|_{\Omega}\in\R^{|\Omega|}$ be entries of $x$ in the index set $\Omega\in [d]$. Let $\Omega^{C} = [d]-\Omega$ be the complement of set $\Omega$.

\vspace{1em}
\begin{definition}[\cite{BCDH10}]
A structured sparsity model $\mathcal{M}_K$ is defined as the union of $m_K$ canonical $K$-dimensional subspaces
\begin{equation}
\mathcal{M}_K = \bigcup_{m=1}^{m_K} \mathcal{X}_m \quad \textnormal{s.t. } \mathcal{X}_m = \{x: x|_{\Omega_m}\in\R^{K}, x|_{\Omega_m^C}=0\},
\end{equation}
where $\{\Omega_1,...,\Omega_{m_K}\}$ is the set containing all allowed supports, with $|\Omega_m|=K$ for each $m = 1, . . ., m_K$, and each
subspace $\mathcal{X}_m$ contains all signals $x$ with $\supp(x)\subset \Omega_m$.

We define $\mathbb{M}(x, K)$ as the algorithm that obtains the best $K$-term structured sparse approximation of $x$ in the union of
subspaces $\mathcal{M}_K$:
\begin{equation}
\mathbb{M}(x, K) = \arg\;\min_{\bar{x}\in \mathcal{M}_K} \norm{x-\bar{x}}_2.
\end{equation}
\end{definition}

We next define an enlarged set of subspaces $\mathcal{M}_K^B$ and the associated sparse approximation algorithm.
\vspace{1em}
\begin{definition}[\cite{BCDH10}]
The $B$-order sum for the set $\mathcal{M}_K$, with $B>1$ an integer, is defined as
\begin{equation}
\mathcal{M}^B_K = \left\{ \sum_{r=1}^B x^{(r)},\quad \text{with } x^{(r)}\in \mathcal{M}_K \right\}.
\end{equation}

We define $\mathbb{M}_B(x, K)$ as the algorithm that obtains the best approximation of $x$ in the union of subspaces $\mathcal{M}_K^B$:
\begin{equation}
\mathbb{M}_B(x, K) = \arg\;\min_{\bar{x}\in \mathcal{M}_K^B} \norm{x-\bar{x}}_2.
\end{equation}
\end{definition}

Algorithm~\ref{cosamp-pos} presents the model-based CoSaMP with positivity constraint. Comparing Algorithm~\ref{cosamp-pos} with the original model-based CoSaMP algorithm (Algorithm 1 of~\cite{BCDH10}), we note that the only different is that Algorithm~\ref{cosamp-pos} has an extra step (Step 6). In Step 6 we take a ReLU operation on $b$ to ensure that $\hat{x}_i$ is always nonnegative after Step 7. 

\begin{algorithm*}[t]
\caption{Model-based CoSaMP with positivity constraint}
\label{cosamp-pos}
\begin{tabbing}
Inputs: measurement matrix $A$, measurements $y$, structured sparse 
approximation algorithm $\mathbb{M}$ \\
Output: $K$-sparse approximation $\hat{x}$ to the true signal $x$, which is assumed to be nonnegative \\
$\hat{x}_0=0$ , $r = y$; $i = 0$ \hspace{37mm} \{initialize\} \\
{\bf while} \= halting criterion false {\bf do} \hspace{20mm}\= \\
\> 1. $i \leftarrow i+1$ \\
\> 2. $e \leftarrow A^T r$ \> \{form signal residual estimate\} \\
\> 3. $\Omega \leftarrow \supp(\mathbb{M}_2(e,K))$ \> \{prune residual estimate according to structure\} \\
\> 4. $T \leftarrow \Omega \cup \supp(\hat{x}_{i-1})$ \> \{merge supports\} \\
\> 5. $b|_T \leftarrow A_T^{\dagger} y$,  $b|_{T^C}\leftarrow 0$ \> \{form signal estimate by least-squares\} \\
\> 6. $\hat{b} = \max\{0, b\}$ \> \{set the negative entries to be zero\} \\ 
\> 7. $\hat{x}_i \leftarrow \mathbb{M}(\hat{b},K)$ \> \{prune signal estimate according to structure\} \\
\> 8. $r \leftarrow y - A \hat{x}_i$ \> \{update measurement residual\} \\
{\bf end while} \\
return $\hat{x} \leftarrow \hat{x}_i$
\end{tabbing}
\end{algorithm*}

We now show that Algorithm~\ref{cosamp-pos} has the same performance guarantee as the original model-based CoSaMP algorithm for structured sparse signals. Speficially, we will show that Theorem 4 of~\cite{BCDH10} also applies to Algorithm~\ref{cosamp-pos}. In~\cite{BCDH10}, the proof of Theorem 4 is based on six lemmas (Appendix D), among which the only lemma that is related to Step 6-7 is Lemma 6. It then suffices to prove that this lemma is also true for Algorithm~\ref{cosamp-pos} under the constraint that the true vector $x$ is nonnegative.

\vspace{1em}
\begin{app-lemma}[Prunning]
The pruned approximation $\hat{x}_i = \mathbb{M}(\hat{b}, K)$ is such that
\begin{equation}
\norm{x-\hat{x}_i}_2 \le 2\norm{x-b}_2.
\end{equation}
\end{app-lemma}
\begin{proof}
Since $\hat{x}_i$ is the $K$-best approximation of $\hat{b}$ in $\mathcal{M}_K$, and $x\in \mathcal{M}_K$, we have
\begin{equation}
\norm{x-\hat{x}_i}_2 \le \norm{x-\hat{b}}_2 + \norm{\hat{b}-\hat{x}_i}_2 \le 2\norm{x-\hat{b}}_2 \le 2\norm{x-b}_2,
\end{equation}
where the last inequality follows from that $\hat{b} = \max\{0, b\}$, and $x\ge0$.
\end{proof}

The above lemma matches Lemma 6, which is used to prove Theorem 4 in~\cite{BCDH10}. Since the other lemmas (i.e., Lemma 1-5 in Appendix D of~\cite{BCDH10}) still hold for Algorithm~\ref{cosamp-pos}, we conclude that the performance guarantee for structured sparse signals (i.e., Theorem 4 of~\cite{BCDH10}) is also true for Algorithm~\ref{cosamp-pos}.

\begin{figure}[ht]
\centering
\includegraphics[width=0.48\textwidth]{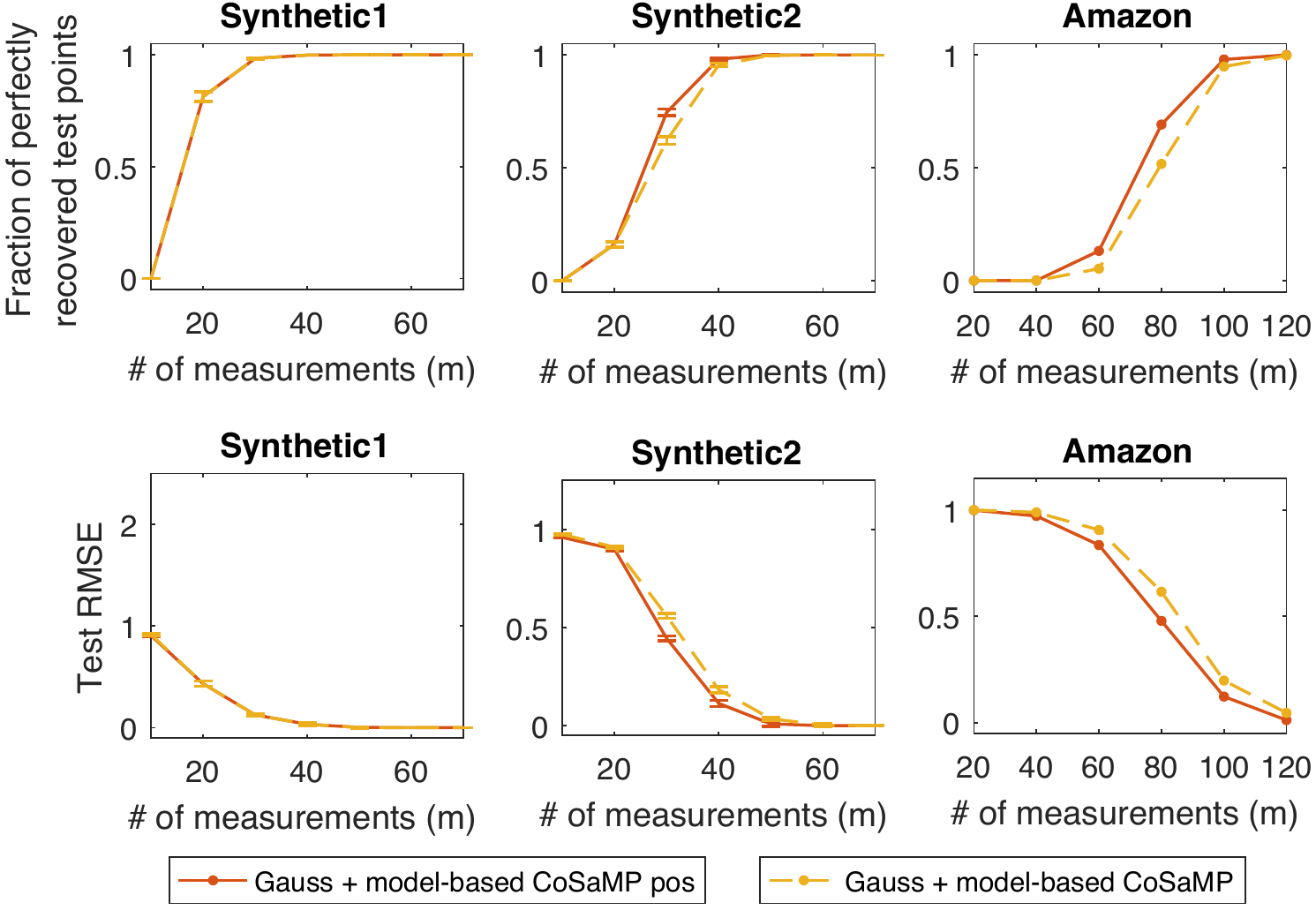}
\vspace{-0.3cm}
\caption{Incorporating the positivity constraint to the model-based CoSaMP algorithm improves its recovery performance.}
\label{cosamp_pos_plot}
\end{figure}

In Figure~\ref{cosamp_pos_plot}, we compare the recovery performance of two decoding algorithms: 1) model-based CoSaMP algorithm (Algorithm 1 of~\cite{BCDH10}) and 2) model-based CoSaMP algorithm with positivity constraint (indicated by ``Model-based CoSaMP pos'' in Figure~\ref{cosamp_pos_plot}). We use random Gaussian matrices as the measurement matrices. Since our sparse datasets are all nonnegative, adding the positivity constraint to the decoding algorithm is able to improve the recovery performance.

\section{Additional experimental results}\label{app-extra-exp}

\subsection{A toy experiment}\label{app-toy}
\begin{figure}[ht]
\centering
\includegraphics[width=0.45\textwidth]{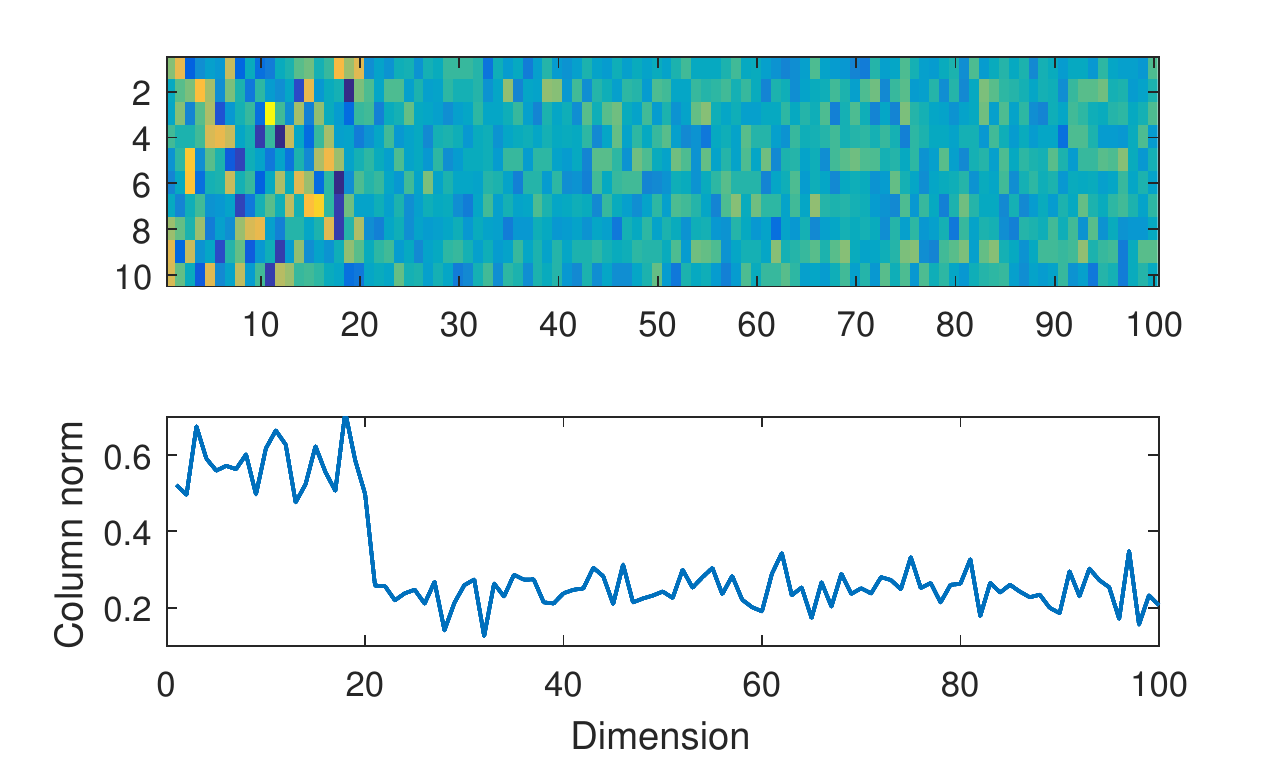}
\vspace{-0.3cm}
\caption{Visualization of the learned matrix $A\in\R^{10\times 100}$ on the toy dataset: a color map of the matrix (upper), the column-wise $\ell_2$ norm (lower). Every sample in the toy dataset has 5 nonzeros, located randomly in the first 20 dimensions.}
\label{toy_plot}
\end{figure}

We use a simple example to illustrate that the measurement matrix learned from our autoencoder is adapted to the training samples. The toy dataset is generated as follows: each vector $x\in\R^{100}$ has 5 nonzeros randomly located in the first 20 dimensions; the nonzeros are random values between [0,1]. We train \ourAE on a training set with 6000 samples. The parameters are $T=10$, $m=10$, and learning rate 0.01. A validation set with 2000 samples is used for early-stopping.After training, we plot the matrix $A$ in Figure~\ref{toy_plot}. The entries with large values are concentrated in the first 20 dimensions. This agrees with the specific structure in the toy dataset.

\subsection{Random partial Fourier matrices}\label{app-extra-exp-fourier}
\begin{figure*}[ht]
\centering
\includegraphics[width=0.75\textwidth]{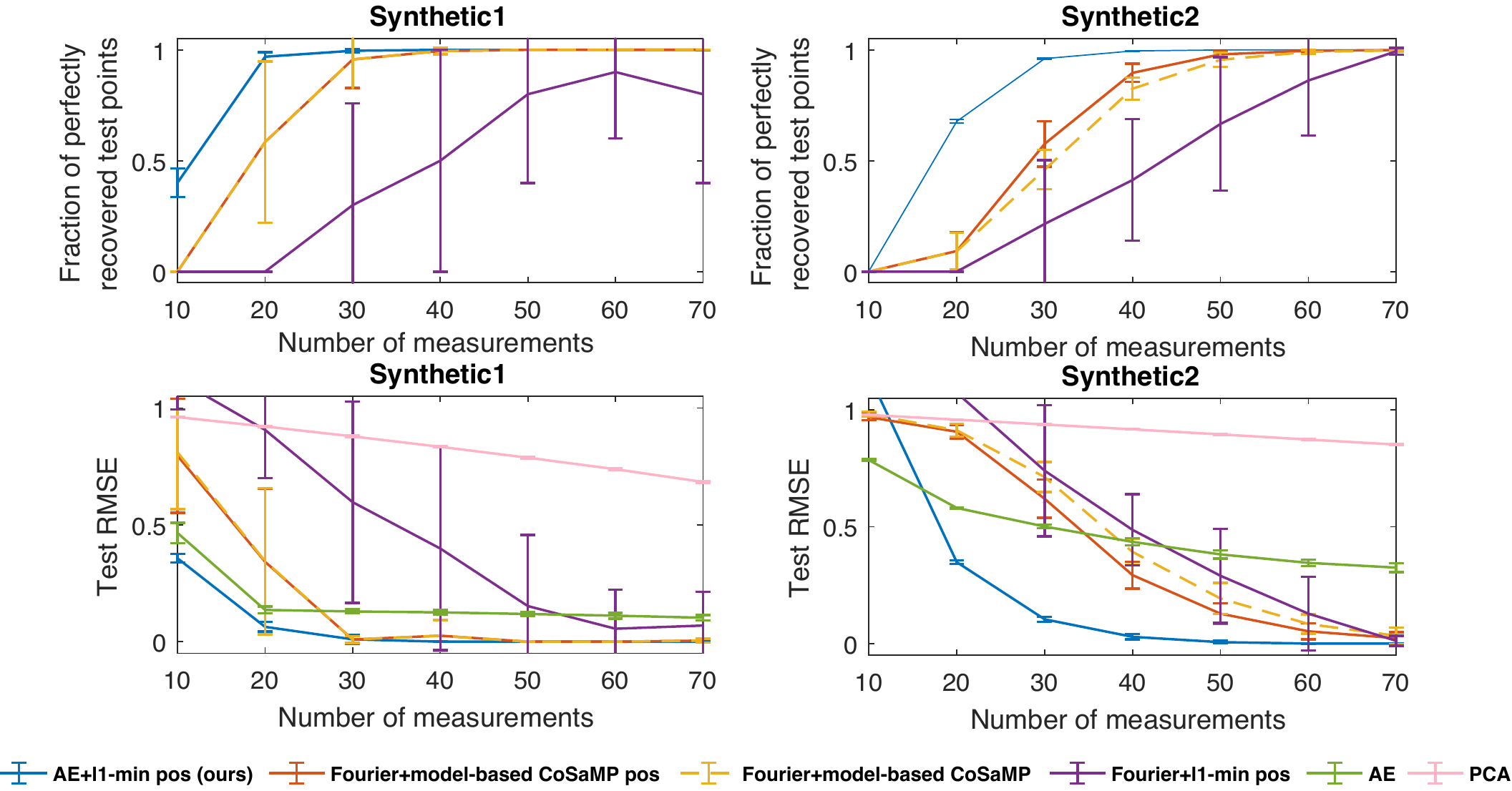}
\caption{Recovery performance of random partial Fourier matrices. Best viewed in color. Similar to Figure~\ref{main_plots}, the error bars represent the standard deviation across 10 randomly generated datasets. We see that the recovery performance of a random partial Fourier matrix (shown in this figure) has a larger variance than that of a random Gaussian matrix (shown in Figure~\ref{main_plots}).}
\label{fourier-plots}
\end{figure*}

Figure~\ref{fourier-plots} is a counterpart of Figure~\ref{main_plots}. The only difference is that in Figure~\ref{fourier-plots} we use random partial Fourier matrices in place of random Gaussian matrices. A random $M\times N$ partial Fourier matrix is obtained by choosing $M$ rows uniformly and independently with replacement from the $N \times N$ discrete Fourier transform (DFT) matrix. We then scale each entry to have absolute value $1/\sqrt{M}$~\cite{HR15}. Because the DFT matrix is complex, to obtain $m$ {\em real} measurements, we draw $m/2$ random rows from a DFT matrix to form the partial Fourier matrix.

A random partial Fourier matrix is a Vandermonde matrix. According to~\cite{DT05}, one can exactly recover a $k$-sparse nonnegative vector from $2k$ measurements using a Vandermonde matrix~\cite{DT05}. However, the Vandermonde matrices are numerically unstable in practice~\cite{Pan16}, which is consistent with our empirical observation. Comparing Figure~\ref{fourier-plots} with Figure~\ref{main_plots}, we see that the recovery performance of a random partial Fourier matrix has larger variance than that of a random Gaussian matrix. 

\subsection{Precision score comparisons for extreme multi-label learning}\label{app-xml}
Table~\ref{app-xml-table} compares the precision scores (P@1, P@3, P@5) over two benchmark datasets. For SLEEC, the precision scores we obtained by running their code (and combining 5 models in the ensemble) are consistent with those reported in the benchmark website~\cite{BDJPV17}. Compared to SLEEC, our method (which learns label embeddings via training an autoencoder \ourAE) is able to achieve better or comparable precision scores. For our method, we have experimented with three prediction approaches (denoted as ``\ourAE  1/2/3'' in Table~\ref{app-xml-table}): 1) use the nearest neighbor method (same as SLEEC); 2) use the decoder of the trained \ourAE (which maps from the embedding space to label space); 3) use an average of the label vectors obtained from 1) and 2). As indicated in Table~\ref{app-xml-table}, the third prediction approach performs the best.
\begin{table*}[ht]
\centering
\caption{Comparisons of precision scores: P@1, P@3, P@5.}
\begin{tabular}{|c|c|c|c|c|c|c|}
\hline
Dataset & \multicolumn{3}{|c|}{EURLex-4K} & \multicolumn{3}{|c|}{Wiki10-31K} \\
\hline
\# models in the ensemble & 1&3&5&1&3&5\\
\hline
SLEEC & 0.7600 & 0.7900&0.7944 & 0.8356 & 0.8603 & 0.8600 \\
\ourAE 1 & 0.7655&0.7928&0.7931 & 0.8529&0.8564&0.8597\\
\ourAE 2 & 0.7949&0.8033&0.8070 & 0.8560&0.8579&0.8583\\
\ourAE 3 & {\bf 0.8062}& {\bf 0.8151}&{\bf 0.8136}&{\bf 0.8617}&{\bf 0.8640}&{\bf 0.8630}\\
\hline
\end{tabular}
\begin{tabular}{|c|c|c|c|c|c|c|}
\hline
Dataset & \multicolumn{3}{|c|}{EURLex-4K} & \multicolumn{3}{|c|}{Wiki10-31K} \\
\hline
\# models in the ensenble & 1&3&5&1&3&5\\
\hline
SLEEC & 0.6116 & 0.6403&0.6444 & 0.7046 & 0.7304 & 0.7357\\
\ourAE 1 & 0.6094&0.0.6347&0.6360 & 0.7230&0.7298&0.7323\\
\ourAE 2 & 0.6284&0.6489&0.6575 & 0.7262&0.7293&0.7296\\
\ourAE 3 & {\bf 0.6500}& {\bf 0.6671}&{\bf 0.6693}&{\bf 0.7361}&{\bf 0.7367}& {\bf 0.7373}\\
\hline
\end{tabular}
\begin{tabular}{|c|c|c|c|c|c|c|}
\hline
Dataset & \multicolumn{3}{|c|}{EURLex-4K} & \multicolumn{3}{|c|}{Wiki10-31K} \\
\hline
\# models in the ensemble & 1&3&5&1&3&5\\
\hline
SLEEC & 0.4965 & 0.5214&0.5275 & 0.5979 & 0.6286 & 0.6311 \\
\ourAE 1 & 0.4966&0.5154&0.5209 & 0.6135&0.6198&0.6230\\
\ourAE 2 & 0.5053&0.5315&0.5421 & 0.6175&0.6245&0.6268\\
\ourAE 3 & {\bf 0.5353}& {\bf 0.5515}&{\bf 0.5549}&{\bf 0.6290}&{\bf 0.6322}&{\bf 0.6341}\\
\hline
\end{tabular}
\label{app-xml-table}
\end{table*}

\subsection{$\ell_1$-minimization with positivity constraint}\label{app-extra-exp-pos-l1}

\begin{figure}[ht]
\centering
\includegraphics[width=0.48\textwidth]{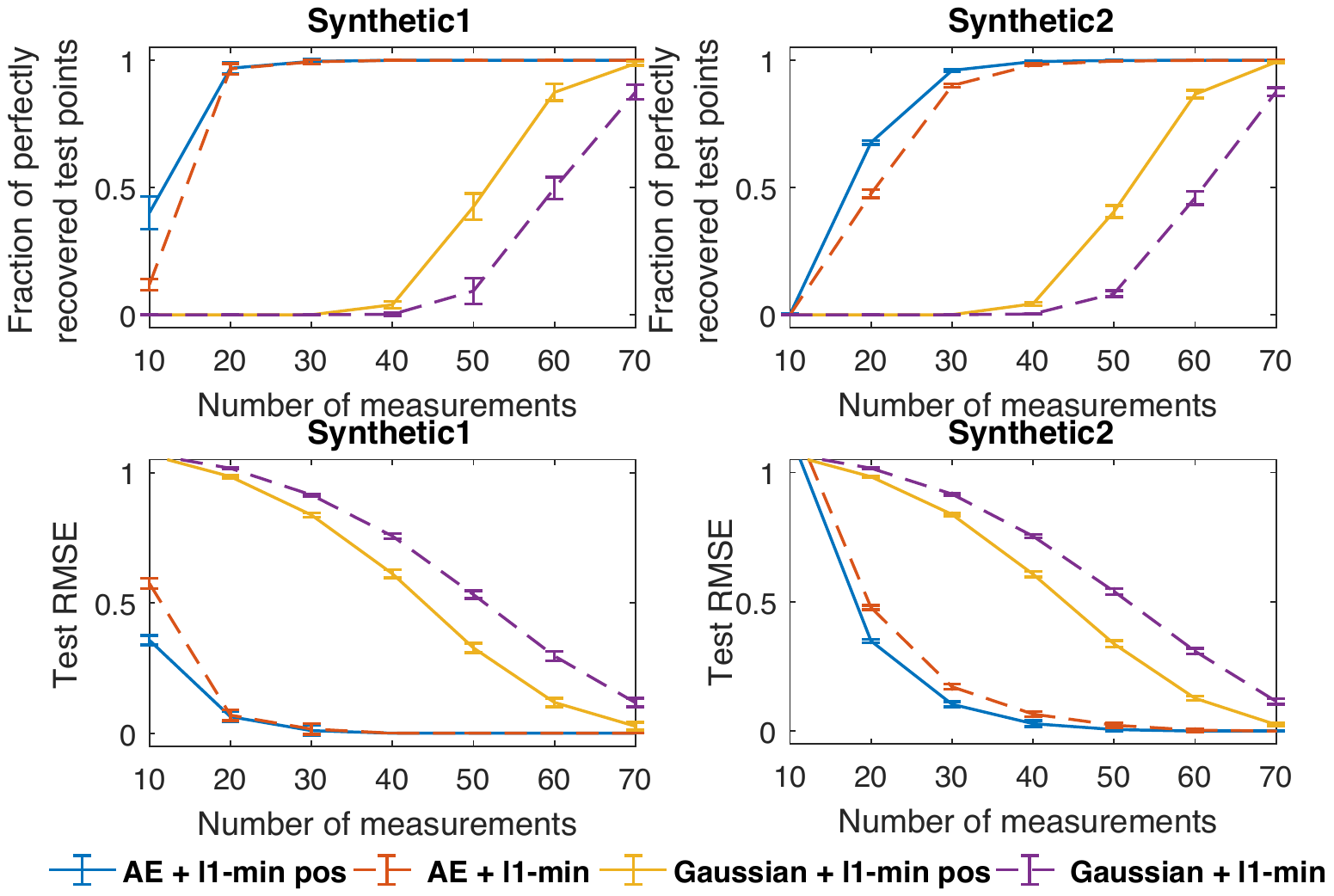}
\caption{A comparison of the recovery performance between $\ell_1$-min (\ref{intuition-l1min}) and the $\ell_1$-min with positivity constraint (\ref{l1-pos}). The sparse recovery performance is measured on the test set. Best viewed in color. We plot the mean and standard deviation (indicated by the error bars) across 10 randomly generated datasets. Adding a positivity constraint to the $\ell_1$-minimization gives better recovery performance than a vanilla $\ell_1$-minimization.}
\label{pos-l1-plots}
\end{figure}

We compare the recovery performance between solving an $\ell_1$-min (\ref{intuition-l1min}) and an $\ell_1$-min with positivity constraint (\ref{l1-pos}). The results are shown in Figure~\ref{pos-l1-plots}. We experiment with two measurement matrices: 1) the one obtained from training our autoencoder, and 2) random Gaussian matrices. As shown in Figure~\ref{pos-l1-plots}, adding a positivity constraint to the $\ell_1$-minimization improves the recovery performance for nonnegative input vectors. 

\subsection{Singular values of the learned measurement matrices}\label{app-singular}

We have shown that the measurement matrix obtained from training our autoencoder is able to capture the sparsity structure of the training data. We are now interested in looking at those data-dependent measurement matrices more closely. Table~\ref{spectrum} shows that those matrices have singular values close to one. Recall that in Section~\ref{sec-intuition} we show that matrices with all singular values being ones have a simple form for the projected subgradient update (\ref{lemma2-update}). Our decoder is designed based on this simple update rule. Although we do not explicitly enforce this constraint during training, Table~\ref{spectrum} indicates that the learned matrices are not far from the constraint set.

\begin{table}[ht]
\centering
\begin{tabular}{|c|c|c|}
\hline
Dataset & $\sigma_{\text{largest}}$ & $\sigma_{\text{smallest}}$\\
\hline
Synthetic1 & 1.117 $\pm$ 0.003 & 0.789 $\pm$ 0.214 \\
Synthetic2 &  1.113 $\pm$ 0.006 &  0.929 $\pm$ 0.259\\
Synthetic3 & 1.162 $\pm$ 0.014 & 0.927 $\pm$ 0.141\\ 
Amazon &  1.040 $\pm$ 0.021 & 0.804 $\pm$ 0.039\\
Wiki10-31K & 1.097 $\pm$ 0.003 & 0.899 $\pm$ 0.044 \\
RCV1 & 1.063 $\pm$ 0.016 & 0.784 $\pm$ 0.034 \\
\hline
\end{tabular}
\caption{Range of the singular values of the measurement matrices $A\in\R^{m\times d}$ obtained from training \ourAE. The mean and standard deviation is computed by varying the number of $m$ (i.e., the ``number of measurements" in Figure~\ref{main_plots}).}
\label{spectrum}
\end{table}

\subsection{Synthetic data with no extra structure}\label{app-no-extra}
We conducted an experiment on a synthetic dataset with no extra structure. Every sample has dimension 1000 and 10 non-zeros, the support of which is randomly selected from all possible support sets. The training/validation/test set contains 6000/2000/2000 random vectors. As shown in Table~\ref{table:no-extra}, the learned measurement matrix has similar performance as a random Gaussian measurement matrix. 
\begin{table}[ht]
\centering
\begin{tabular}{|c|c|c|c|}
\hline
\# measurements & 30 & 50  & 70\\
\hline
\ourAE + $\ell_1$-min pos &0.8084&0.1901&0.0016\\
\hline
Gaussian + $\ell_1$-min pos &0.8187&0.1955&0.0003\\
\hline
\end{tabular}
\caption{Comparison of test RMSE on a synthetic dataset with no extra structure.}
\label{table:no-extra}
\end{table}

\subsection{Autoencoder with unrolled ISTA}\label{app-LISTA-AE}

Our autoencoder \ourAE is designed by unrolling the projected subgradient algorithm of the standard $\ell_1$-minimization decoder. We can indeed design a different autoencoder by unrolling other algorithms. One option is the ISTA (Iterative Shrinkage-Thresholding Algorithm) algorithm of the LASSO problem. Comparing the performance of those autoencoders is definitely an interesting direction for future work. We have performed some initial experiments on this. We designed a new autoencoder ISTA-AE with a linear encoder and a nonlinear decoder by unrolling the ISTA algrithm. On the Synthetic1 dataset, \ourAE performed better than LISTA (we unrolled ten steps for both decoders): with 10 measurements, the test RMSEs are 0.894 (ISTA-AE), 0.795 (ISTA-AE + $\ell_1$-min pos), 0.465 (\ourAE) and 0.357 (\ourAE + $\ell_1$-min pos).

\subsection{Additional experiments of LBCS} \label{app-LBCS}
\begin{figure}[ht]
\centering
\includegraphics[width=0.49\textwidth]{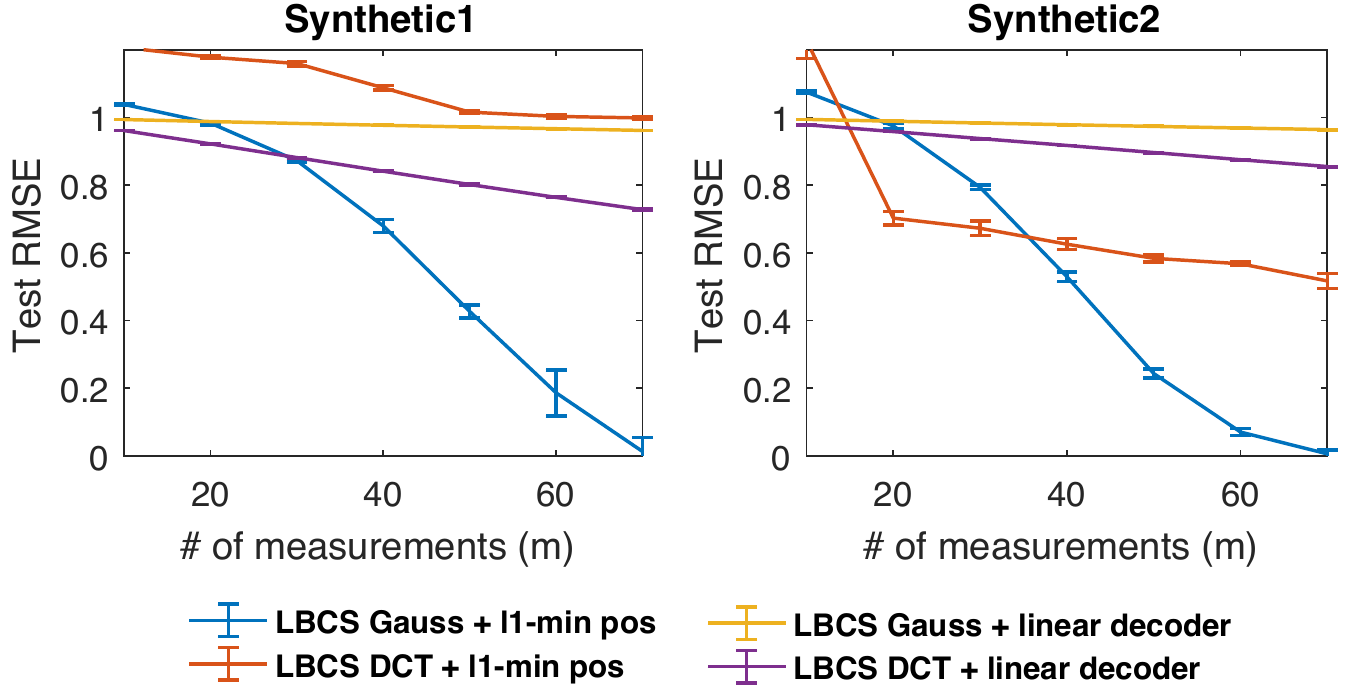}
\caption{We compare four variations of the LBCS method proposed in~\cite{BLSGBC16, LC16}: two basis matrices (random Gaussian and DCT matrix); two decoders ($\ell_1$-minimization and linear decoding). The combination of ``Gaussian + $\ell_1$-minimization'' performs the best. Best viewed in color. For each method, we plot the mean and standard deviation (indicated by the error bars) across 10 randomly generated datasets.}
\label{LBCS_plot}
\end{figure}

We experimented with four variations of LBCS: two different basis matrices (random Gaussian matrix and DCT matrix), two different decoders ($\ell_1$-minimization and linear decoder). As shown in Figure~\ref{LBCS_plot}, the combination of Gaussian and $\ell_1$-minimization performs the best.

\end{document}